\documentclass[conference]{IEEEtran}
\ifCLASSINFOpdf
\else
\fi
\usepackage{cite}
\usepackage{amsmath,amsfonts,amssymb}
\usepackage{algorithmic}
\usepackage{graphicx}
\usepackage{rotating,makecell}
\usepackage{textcomp}
\usepackage{xcolor}
\usepackage{url}

\usepackage{xspace}
\usepackage{multirow}

\usepackage[english,ruled,lined,linesnumbered]{algorithm2e}
\usepackage{algorithmic}

\usepackage{amssymb}
\usepackage{pifont}

\usepackage{tabularx}
\usepackage{colortbl}

\newcommand{\X}{\textsc{LimeOut}\xspace} 

\newcommand{\F}{\textsc{FixOut}\xspace} 

\newcommand{\FK}{\textsc{Find-K}\xspace}

\begin{document}
%

\title{Reducing Unintended Bias of ML Models on Tabular and Textual Data}



%
\author{\IEEEauthorblockN{Guilherme Alves,
Maxime Amblard,
Fabien Bernier,
Miguel Couceiro  and 
Amedeo Napoli}
\IEEEauthorblockA{
Université de Lorraine, CNRS, Inria Nancy G.E., LORIA \\
Vandoeuvre-lès-Nancy, France \\
\{guilherme.alves-da-silva, maxime.amblard, fabien.bernier, miguel.couceiro, amedeo.napoli\}@loria.fr
}
}


\maketitle

\begin{abstract}
Unintended biases in machine learning (ML) models are among the major concerns that must be addressed to maintain public trust in ML.
In this paper, we address \textit{process fairness} of ML models
that consists in reducing the dependence of  models' on sensitive features, without compromising their performance.
We revisit the framework \F that is inspired in the approach ``fairness through unawareness'' to build fairer models. We introduce several improvements such as automating  the choice of \F's parameters. Also, \F was originally proposed to improve 
fairness of ML models on tabular data. We also demonstrate the feasibility of \F's workflow for models on textual data. We present several experimental results that illustrate the fact that \F improves process fairness on different classification settings.

\end{abstract}


%
\IEEEpeerreviewmaketitle

\section{Introduction}\label{sec:intro}


The pervasiveness of machine learning (ML) models in daily life has raised several issues. One is the need for public trust in algorithmic outcomes. In 2016, the European Union (EU) enforced the General Data Protection Regulation (GDPR)\footnote{\url{https://gdpr-info.eu/}} that gives users the right to understand the inner workings of ML models and to obtain explanations of their outcomes, which can increase public trust in ML outcomes.

These explanations can also reveal biases of algorithmic decisions. ML models are designed to have some bias that guide them in their target tasks. For instance, a model for credit card default prediction is intended to be biased so that users who have a good credit payment history get a higher score than those who have not. Similarly, a model for hate speech detection is intended to be biased such that messages that contain offensive terms get higher scores than those which have not. However, the model's outcomes should not rely on salient nor discriminatory features that carry prejudice towards unprivileged groups \cite{grgic2016case}. For instance, the model that predicts credit card defaults should not give lower scores to users because they belong to a particular ethnicity compared to those who do not. In the same way, the model to detect hate speech should not give higher scores to messages that are written in a particular language variant compared to those messages that are not. These are known as unintended biases \cite{dixon2018measuring}.

Unintended bias leads to various consequences. If it negatively impacts society, in particular against minorities, then it characterizes an unfairness issue~\cite{dixon2018measuring,hardt2016equality}.
Researchers found unintended bias and fairness issues in a wide range of domains and data types.
For instance, it was found racial bias in the system that predicts criminal recidivism, COMPAS\footnote{\url{https://www.propublica.org/article/machine-bias-risk-assessments-in-\ criminal-sentencing}}.
Gender and ethnic discriminations were also detected in online platforms for job applications~\cite{hangartner2021monitoring}.
Bias based on language variants was found on approaches that detect abusive language, in which messages written in African-American English got higher toxicity scores than messages in Standard-American English~\cite{davidson2019racial}.

In order to address algorithmic fairness, various notions of fairness have been defined based on models' outcomes \cite{catuscia}. 
One relies on the fairness metrics based on standard classification metrics computed on different sub-populations.
Another notion is process fairness which, instead of focusing on the final outcomes, is centred on the process that leads to the outcomes (classifications)~\cite{grgic2016case}.
One way to achieve the latter is to seek \textit{fairness through unawareness}, i.e., to avoid the use of any sensitive features when classifying instances~\cite{kusner2017counterfactual}.
However, dropping out sensitive features may compromise classification performance~\cite{zafar2017fairness}.

Recently, two frameworks have been proposed to address algorithmic unfairness of ML models on tabular data without compromising its classification performance. \X~\cite{bhargava} and \F \cite{alves} render pre-trained models fairer by decreasing their dependence on sensitive features. For that, users must input their pre-trained models together with the features they consider sensitive. \X and \F then use explanation methods to assess fairness of the pre-trained models. In the case a pre-trained model is deemed unfair, \F applies a \textit{feature dropout} followed by an ensemble approach to build a model that is fairer compared to the pre-trained model. Both \X and \F use model-agnostic explanation methods to assess fairness. However, \X only relies on LIME explanations~\cite{ribeiro} where as \F is able to apply any explanation method based on feature importance. In addition, they require beforehand users to define the sampling size of instances and the number of features used to assess fairness of pre-trained models, which raises the need for automation or, at least, fine-tuning. 

These frameworks were initially proposed to address algorithmic unfairness on tabular data. The empirical studies have shown that both frameworks improve process fairness without compromising accuracy. However, fairness issues have also been found in textual data.
For instance, Papakyriakopoulos et al.~\cite{papakyriakopoulos2020bias} trained word embeddings on texts from Wikipedia articles and from political social media (tweets and Facebook comments) that are written in gendered languages\footnote{The grammar of gendered languages require that speakers mark gender in some parts of speech, for instance in nouns and/or adjectives (e.g. German and French).}. They then investigated the diffusion of bias when these embeddings are used in ML models. In addition, they proposed approaches to reduce bias in embeddings based on gender, ethnicity and sexual orientation. In spite of eliminating bias in word embeddings, the authors state that it might include other types of bias.

In this paper, we take advantage of the existing approaches that address algorithmic unfairness through unawareness, and extend them to other settings, namely classification scenarios dealing with textual data. Our main hypothesis is that \textit{\F is able to improve process fairness not only in tabular data but also in textual data}. It is addressed by investigating two research questions:

\begin{itemize}
    \item[Q1.] How to reduce human intervention in \F usage on tabular data? 
    \item[Q2.] Does \F reduce model's dependence on sensitive words on textual data?
\end{itemize}

The main contributions of the work described herein can be summarized as follows: (1) an approach that automatically selects the number of features used by \F to assess fairness, and (2) an instantiation of \F that is able to improve process fairness on textual data.


This paper is organized as follows.
We first recall related work about fairness and explanations in Section \ref{sec:literature}.
We then describe \F, our automation of \F on tabular data and the instantiation of \F to textual data in Sections~\ref{sec:fixout}, ~\ref{sec:auto_k}, and~\ref{sec:desc_text_data}, respectively. The experimental setup and results are driven by questions Q1 and Q2 and discussed in Section \ref{sec:experiments}. We conclude this paper in Section \ref{sec:conclusion} together with the future works.

\section{Related Work}\label{sec:literature}

In this section, we introduce the main concepts underlying this work.
We start by recalling the basics of fairness notions in Section \ref{sec:fairness_metrics}. We then briefly present the fundamentals of some explanation methods in Section \ref{sec:exp_methods}. 

\subsection{Assessing Fairness in ML Models} \label{sec:fairness_metrics}

Several approaches have been proposed to assess fairness in ML models; most of them focused on the algorithmic outcomes \cite{hardt2016equality}. They can be divided into two main families of fairness notions \cite{catuscia}, as we briefly outline next.


\begin{itemize}
    \item \textit{Group fairness} considers that groups differing by sensitive features (and ignoring all non-sensitive features) must be treated equally \cite{catuscia}. For instance, demographic parity \cite{hardt2016equality}, disparate mistreatment \cite{GummadiDisparateMistreatment}, and predictive equality \cite{catuscia} are metrics that belong to this family of fairness notions.
    \item \textit{Individual fairness} states that similar individuals (non-sensitive features included) should obtain similar probabilities of getting similar outcomes\cite{speicher2018unified}.
    For instance, counterfactual fairness metrics \cite{wachter2018counterfactual} belong to this family of fairness notions.
\end{itemize}

In this work, we focus on another type of approach to assess fairness, which is called \textit{process fairness} \cite{grgic2016case,bhargava}. 
Process fairness is centered on the process itself rather that focusing on the final outcome.
The idea is to monitor input features used by ML models for ensuring that these ML models do not use sensitive features while making classifications.

\subsection{Explanation Methods} \label{sec:exp_methods}

Explanation methods can be divided into two main groups: (1) those that provide local explanations and (2) those that compute global explanations \cite{guidotti2018survey}.
Local explanation methods generate explanations for individual predictions, while global explanations give an understanding of the global behaviour of the model.

The first group can be divided w.r.t. the type of explanations: methods based on feature importance \cite{ribeiro,shap}, saliency maps \cite{adebayo2018sanity}, and counterfactual methods \cite{wachter2018counterfactual}. In the same way, global explanation methods can be divided into methods based on a collection of local explanations \cite{ribeiro}, and representation based explanations \cite{kim2018interpretability}.
In this work, we are interested in model agnostic explanation methods and thus we focus on LIME and SHAP.

LIME and SHAP provide explanations that are obtained after approximating feature values and predictions.
Moreover, LIME and SHAP are model-agnostic, which means that both can explain any prediction model (classifier).

Let $f$ be the prediction model, $x$ be a target data instance, and $f(x)$ be the prediction we want to explain.
In order to explain $f(x)$, LIME and SHAP need to have access to a neighbourhood of $x$.
Then they generate data instances around $x$ by applying perturbations. While generating neighbors of $x$, these methods try to ensure interpretability by using interpretable versions of $x$ (denoted as $x'$) and of its neighbourhood. 
A mapping function $h_x(x')$ is responsible for converting $x'$ from the interpretable space to the feature space.
For instance, different data types require distinct mapping functions $h_x$.
For tabular data, $h_x$ treats discretized versions of numerical features, while for textual data, it deals with the presence/absence of words. One example of this mapping is depicted in Figure \ref{fig:map_func}.

Now, let $\xi$ be the explanation for $f(x)$.
Explanations take the form of surrogate models that are linear models (transparent by design) for LIME and SHAP.
They learn a linear function $g$, i.e., $g(z')=w_g \cdot z'$, where $w_g$ are the weights of the models which correspond to the importance of features.
The function $g$ should optimize the following objective function:

\begin{equation}
 \label{eq:obj_func_exp}
 \xi = \arg\min_{g \in \mathcal{G}} \left \{ L(f,g,\pi_x) + \Omega(g)  \right \},
\end{equation} 

\noindent
$\Omega$ measures the complexity in order to insure interpretability of the linear model given by the explainer.
$L$ is the loss function defined by:

\begin{equation}
 \label{eq:loss_func}
 L(f,g,\pi_x) = \sum_{z,z' \in Z} [f(z)-g(z')]^2 \pi_x(z),
\end{equation}

\noindent where $z$ is the interpretable representation of $x$, and $\pi_x(z)$ defines the neighborhood of $x$ that is considered to explain $f(x)$.

In spite of sharing some commonalities, LIME and SHAP differs in the definition of the kernel $\pi_x$ and also in the complexity function $\Omega$ used to produce explanations.
In the following we recall each method and highlight their differences.

\subsubsection{LIME}
\textbf{L}ocal \textbf{I}nterpretable \textbf{M}odel Agostic \textbf{E}xplanations
is a model-agnostic explanation method providing local explanations \cite{ribeiro,garreau2020explaining}.
Explanations obtained from LIME take the form of surrogate linear models. LIME learns a linear model by approximating the prediction and feature values in order to mimic the behavior of ML model.
To do so, LIME uses the following RBF kernel $\pi_x$ to define the neighborhood of $x$ and considered to explain $f(x)$:

\[
    \pi_x(z)=exp(-\delta(x,z)^2/\sigma^2),
\]
\noindent where $\delta$ is a distance function between $x$ and $z$, and $\sigma$ is the kernel-width. LIME tries to minimize $\Omega(g)$ for reducing the number of non-zero coefficients in the linear model. and $K$ is the hyper parameter that limit the desired complexity of explanations.  


\subsubsection{SHAP}

\begin{figure}
    \centering
    \includegraphics[width=1\columnwidth]{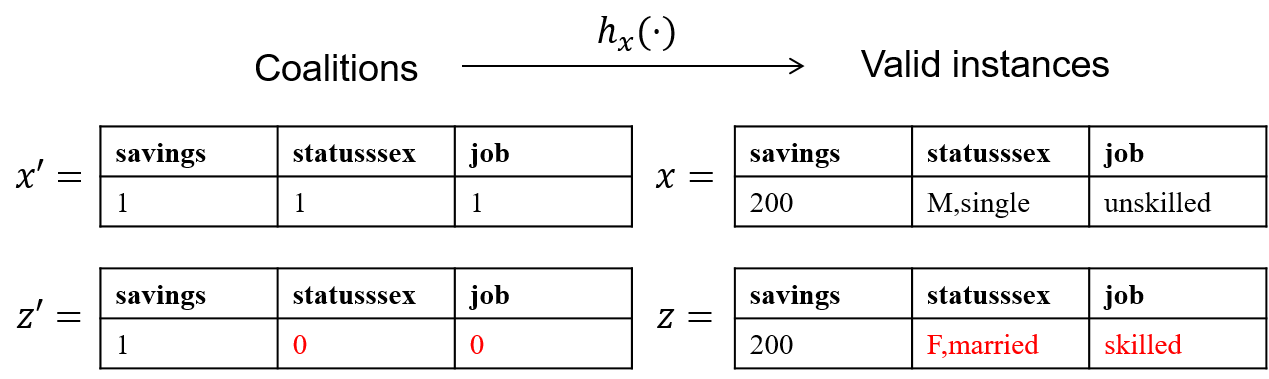}
    \caption{An example of a mapping function $h_x$ that converts  coalitions to valid instances (i.e., instances in the feature space).}
    \label{fig:map_func}
\end{figure}

\textbf{SH}apley \textbf{A}dditive ex\textbf{P}lanations is also an explanation method that provides local explanations. SHAP is based on coalitional game theory. It uses the notion of coalition of features to estimate the contribution of each feature to the prediction $f(x)$. Coalition of features defines the representation space for SHAP and it indicates which features are present in $x$ (see Figure \ref{fig:map_func}). Lundenber et al. \cite{shap} proposed a model-agnostic version of SHAP called KernelSHAP. Like LIME, KernelSHAP also minimizes the loss function described in Eq.~\ref{eq:loss_func}, but it uses a different kernel and, unlike LIME, KernelSHAP sets $\Omega(g)=0$ for the complexity of explanations. 

Let $|z|$ be the number of present features in the coalition $z$, $M$ be the maximum coalition size, and $m$ be the number of coalitions.
The kernel function $\pi_x(z)$ used by KernelSHAP is the following:
\[
    \pi_x(z)=\frac{M-1}{{M \choose |z|}|z|(M-|z|)}.
\]




Both LIME and SHAP only provide explanations for individual (local) predictions. However, in order to detect unintended bias, a global understanding of the inner workings of the ML model is needed. In the following subsection we recall a simple strategy to obtain global explanations from individual explanations.

\subsubsection{From Local to Global Explanations}

Local explanation methods have been used to explain individual predictions.
However, they only provide local explanations which alone cannot provide a global understanding of ML models.
To overcome this issue, Ribeiro et al.~\cite{ribeiro} proposed the so-called Submodular-pick (SP) for selecting a subset of local explanations that helps in interpreting the global behavior of any ML model.
SP was originally proposed to work along with LIME explanations and it is called SP-LIME.
The main idea on which relies SP-LIME is to sample a set of instances whose explanations are not redundant and that has a ``high covering'' in the following sense.

Denote by $B$ the desired number of explanations used to explain $f$ globally, and let $V$ be a set of selected instances, $I$  an array of feature importance, and $W$ an explanation matrix --columns represent features and rows represent instances-- that contains the importance (contribution) of $d'$ features to each instance.
SP-LIME picks instances that are explaining thanks to:
\[
    Pick(W,I) = \arg\max_{V,|V|<B} \sum_{j=1}^{d'} \mathrm{1}_{[i \in V : \mathcal{W}_{ij} > 0]} I_j.
\]


\section{\F} \label{sec:fixout}

In this section, we revisit the human-centered framework \F that tackles process fairness \cite{alves}.
\F is model-agnostic and explainer-agnostic such that it can be applied on top of any ML model, and in addition it is also able to use any explanation method based on feature importance. 

\F takes as input a quadruple  $(M,D,S,E)$ of a pre-trained model $M$, a dataset $D$ used to train $M$, a set of sensitive features $S$, and an explanation method $E$.
It is based on two main components which are made precise below. 

The first component is in charge of assessing fairness of $M$. More precisely, explainer $E$ builds the list $L$ of features  that contribute the most to the classifications of $M$ (function \textsc{GlobalExplanations} in the Algorithm \ref{algo:fixout}). Here, users must indicate a value for the parameter $k$ that limits the size of this list.
\F then applies the following rule: if at least one sensitive feature returned by $E$ appears in $L$, $M$ is deemed unfair.
Otherwise, $M$ is deemed fair and no action is taken.

The second component is applied only if $M$ is deemed unfair. The goal of this component is to use feature dropout together with an ensemble approach to build an ensemble classifier $M_{final}$ which is less dependent on sensitive features --or fairer-- than $M$ (see Algorithm \ref{algo:fixout}). 

Let $S' \subseteq S$ be the subset of sensitive features appearing in $L$.
\F uses feature dropout to train $|S'|+1$ classifiers in the following way: $|S'|$ classifiers are defined by removing each of the sensitive features in $S'$, and an additional classifier is defined by removing all sensitive features in $S'$.

Now, let $x$ be an instance, $C$ a class value, and $M_t$ the $t$-th classifier in the pool of classifiers. \F uses models that make classifications in the form of probabilities $P(x \in C)$.
Once all classifiers are trained, they are then combined thanks to an aggregation function. \X uses only a simple average while \F is able to employ either a simple or weighted average function.

The aggregation function is responsible for producing the final classification. The final probability of $x$ to belong to the class $C$ is based on the probabilities $P(x \in C)$ obtained from the pool of classifiers and it is defined as the simple average:
\begin{equation} \label{eq:new_aggregation}
    P_{M_{final}}(x \in C) =  \frac{1}{|S'|+1}\sum_{t=1}^{|S'|+1} P_{M_t}(x \in C),
\end{equation}
\noindent where $P_{M_t}(x \in C)$ is the probability predicted by model $M_t$.


\begin{algorithm}
\KwData{$M$: pre-trained model; $D$: dataset; $S$: sensitive features; $E$: explainer; $k$: number of features considered.}
\KwResult{$M_{final}$ an ensemble model.}

$L \gets \textsc{GlobalExplanations}(E,M,D,k)$ \\
$S' \gets \{ s_t \mid s_t \in S \wedge s_t \in L\}$ \\
\uIf{$|S'| > 0$}{
    $M_{final} \gets \{  \textsc{Train}(M_t,D-\{s_t\}) \mid  s_t \in S' \}$ \\
    $M_{final} \gets M_{final} \cup \textsc{Train}(M_{t+1},D-S')$ \\
    \Return $M_{final}$ \\
}

\caption{\F}
\label{algo:fixout}
\end{algorithm}

\begin{table}
     \centering
     \caption{An example of global explanations for Random Forest on the German dataset.}
     \footnotesize
         \setlength{\tabcolsep}{5pt}
         \begin{tabular}{|l|l|}
             \multicolumn{2}{c}{Original (SHAP)}   \\
             \hline
             \textbf{Feature} & \textbf{Contrib.} \\
             \hline
             \rowcolor{black!10}\textbf{statussex} & -10.758909 \\
             property & 10.676458 \\
             credithistory & 10.264842 \\
             residencesince & 8.108638 \\
             employmentsince & 6.818476 \\
             existingchecking & 6.308758 \\
             housing & -5.649528 \\
             installmentrate & 5.125154 \\
             duration & 4.838629 \\
             \rowcolor{black!10}\textbf{telephone} & 3.981387 \\
             \hline
         \end{tabular}
         \hspace{10pt}
         \begin{tabular}{|l|l|}
             \multicolumn{2}{c}{\F using SHAP}   \\
             \hline
             \textbf{Feature} & \textbf{Contrib.} \\
             \hline
             credithistory & 13.246923 \\
             employmentsince & 9.550589 \\
             property & 8.958718 \\
             residencesince & 7.264848 \\
             installmentrate & 5.893469 \\
             housing & -3.829469 \\
             \rowcolor{black!10}\textbf{statussex} & -3.313667 \\
             duration & 2.678797 \\
             existingcredits & 2.412488 \\
             \rowcolor{black!10}\textbf{telephone} & 2.269493 \\
             \hline
     \end{tabular}
     \label{tb:exp-german}
 \end{table}

\section{Automating the choice of $k$} \label{sec:auto_k}

In order to assess fairness of ML models, \F only takes into account the first $k$ most important features.
Thus, \F practitioners must know beforehand a suitable value for $k$ so that \F builds $L$ with the $k$ most important features.
The system then looks at whether sensitive features are in $L$.
An inappropriate value for $k$ prevents \F of correctly detecting unfairness issues as all sensitive features may not appear in $L$.


In this section, we propose an algorithm that automates the choice of $k$, based on the statistical measure \textit{kurtosis}.
Indeed kurtosis indicates the flatness of a distribution.
Let $X=(x_i)_{i\in\mathbb{N}}$ be a random variable, the \textit{kurtosis} $\gamma$ of $X$ is defined as follows:

\[
    \gamma(X) = \frac{1}{n} \sum_{x_i} \left ( \frac{x_i-\mu}{\sigma} \right ) ^4,
\]

\noindent where $\sigma$ is the standard deviation, $\mu$ is the mean, $x_i$ is a sample of $X$. For instance, $\gamma(X) < 3$ indicates that $X$ is \textit{platykurtic} (flattened), while $\gamma(X) = 3$ defines a \textit{mesokurtic} distribution (which is the case of normal distribution), and  $\gamma(X) > 3$ indicates that the distribution is \textit{leptokurtic} (pointed) \cite{moors1986meaning,saporta2006probabilites}.

Accordingly we propose the iterative algorithm called \FK (see Algorithm \ref{algo:findK}), that selects a sub-sample of features of $L$ based on the kurtosis measure.
\FK removes features from $L$ by analyzing the kurtosis $\gamma(L)$ of $L$ before and after the deletion of a subset of features.
For that, it iteratively removes features from the less important to the most important.
The idea on which relies this algorithm is that once a feature is removed, the kurtosis of $L$ changes.
The algorithm stops when $|\gamma(L)-\gamma(L')| > \alpha$, where $L$ is the original list, $L'$ is the list obtained after removing features, and $\alpha > 0$ is a parameter.

\begin{algorithm}[htpb]
\KwData{$L$: sorted list of contributions of all features (descending order); $\alpha$: a threshold.}
\KwResult{$L'$ a new list of contributions of subset of features.}

$L' \gets L$ \\
\For{$i \gets |L|$ \text{\textbf{to}} $2$}{
    $L' \gets L' - L'[i]$ \\
    \uIf{$|\gamma(L)-\gamma(L')| > \alpha$}{
        \text{\textbf{break}}
    }
}
\caption{\FK}
\label{algo:findK}
\end{algorithm}

\section{\F on Textual Data} \label{sec:desc_text_data}

In this section we demonstrate the adaptability of the \F framework to another data type, namely textual data.
In this setting, a data instance corresponds to a text and features to words, i.e., the number of features is equal to the vocabulary size.
To drop a feature, we remove each occurrence of the corresponding word in the text.
Here words that are semantically equivalent are not taken into account.


Let $M$ be a classifier for which we want to reduce the impact of  words from a given set $S$, in the classifier's outcomes:
\[
S = \{s_i \mid 1 \leqslant i \leqslant n\}.
\]
As in tabular data, \F decreases the contribution of the words in $S$ to the classifiers outcomes by applying feature dropout and by building an ensemble of the $n+1$ classifiers:
\begin{itemize}
    \item one classifier $M_i$, for $1 \leqslant i \leqslant n$, trained by ignoring the word $s_i$, and
    \item a classifier $M_{[n+1]}$ trained by ignoring all words in $S$.
\end{itemize}

However, when $n$ is large, this method becomes inefficient, for two reasons.
The first one is that if we consider a sensitive word $s_i \in S$, only two classifiers are ignoring it: $M_i$ and $M_{[n+1]}$, i.e., $n-1$ out of the $n+1$ classifiers are taking the feature $s_i$ into account.
Hence, the dropout of $s_i$ may then become less significant when $n$ is large.
The second reason is a complexity concern: the more models there are, the more memory and computation time the ensemble takes.

Table~\ref{tab:text_example1} shows satisfactory results for a random forest model, which classifies a text as hate speech or not (from the \textit{hatespeech} database, which is described later).
We select $n =$ 3 words for which we want to reduce the contribution.
The list of words used here was inspired by the analysis performed in \cite{davidson2019racial}.
We train a new model and globally explain it 50 times (with Random Sampling), in order to retrieve an average contribution.
We can check that the rank of importance and contribution of each dropped feature decreases thanks to the \F ensemble.
We evaluate the gain by calculating the difference between the rank obtained and the starting rank, normalised by the smaller of the two. Gaining a few ranks on the highest features in the ranking allows us to have significant scores, whereas they will be close to 0 for ranks that are very far away.
However, when $n$ increases and  more than 3 words are added, the method seems to become less effective, as shown in Table~\ref{tab:text_example2} (without grouping), which describes the rank and contribution of each one of the selected words whose importance should be reduced.

Let us turn our attention to the interest of grouping words.
Instead of ignoring a single word or feature, the classifier can drop many words at the same time, then reducing the size of the ensemble.
We proceeded in this way and the performance of \F was improved as it can be observed in Table~\ref{tab:text_example2}.


\begin{table}[!ht]
    \centering
    \caption{Process fairness assessment on a hate speech classifier (random forest), global explanation with SHAP RS, selecting 3 words}
    \begin{tabular}{|c||c|c||c|c||c|}
        \hline
        & \multicolumn{2}{c||}{{Original model}} & \multicolumn{2}{c||}{{\F Ensemble}} & Diff\\
        \textbf{Word} & \textbf{Rank} & \textbf{Contrib.} & \textbf{Rank} & \textbf{Contrib.} &\textbf{Rank}\\
        \hline
        \textit{niggah} & 14 & 0.176 & 30 & 0.043 & 1.14\\
        \textit{nigger} & 12 & 0.213 & 29 & 0.045 & 1.41\\
        \textit{nig} & 7 & 0.276 & 55 & 0.021 & 6.85\\
        \hline
    \end{tabular}
    \label{tab:text_example1}
\end{table}


\begin{table}[!ht]
    \centering
    \caption{Process fairness assessment on a hate speech classifier (random forest), global explanation with SHAP RS, selecting 7 words}
    \begin{tabular}{|c||c|c||c|c||c|}
        \hline
        & \multicolumn{2}{c||}{{Without grouping}} & \multicolumn{2}{c||}{{With grouping}} & Diff\\
        \textbf{Word} & \textbf{Rank} & \textbf{Contrib.} & \textbf{Rank} & \textbf{Contrib.} & \textbf{Rank}\\
        \hline
        \textit{niggah} & 18 & 0.149 & 23 & 0.03 & 0.27\\
        \textit{nigger} & 15 & 0.164 & 21 & 0.031 & 0.40\\
        \textit{nigguh} & 22 & 0.13 & 83 & 0.008 & 2.77\\
        \textit{nig} & 12 & 0.202 & 65 & 0.011 & 4.41\\
        \textit{nicca} & 22 & 0.107 & 39 & 0.018 & 0.77\\
        \textit{nigga} & 20 & 0.125 & 12 & 0.067 & -0.66\\
        \textit{white} & 25 & 0.087 & 36 & 0.018 & 0.44\\
        \hline
    \end{tabular}
    \label{tab:text_example2}
\end{table}


\section{Experimental Results and Discussion}\label{sec:experiments}
The main goal of the experiments is to demonstrate the adaptability of \F on different data types. This section is thus divided in two main parts, the first one focusing on tabular data and the second one on textual data.
We start by describing the tabular datasets (\S~\ref{sec:desc_text_data}).
We then discuss the empirical results related to the selection of instances for assessing fairness (\S~\ref{sec:exp_sampling_size}), and the automation of the choice of $k$ (\S~\ref{sec:exp_auto_k}).
We end up this section by describing the textual dataset (\S~\ref{sec:exp_text_data}) and by discussing the results obtained from the experiments performed on this data type (\S~\ref{sec:exp_text_proc_fairness}).

\subsection{Tabular Datasets} \label{sec:desc_tab_data}

The experiments on tabular data were performed on 3 real-world datasets, which
 contain sensitive features and are briefly outlined below.


 \textbf{German.}
    This dataset contains 1000 instances and it is available in the UCI repository\footnote{\url{https://archive.ics.uci.edu/ml/datasets/statlog+(german+credit+data}}. For this dataset, the goal is to predict if the credit risk of a person is good or bad.
    Each applicant is described by 20 features in total out of which 3 features, namely ``statussex'', ``telephone'', ``foreign worker'', are considered as sensitive.
    
    \textbf{Adult.}
    This dataset is also available in the UCI repository\footnote{\url{http://archive.ics.uci.edu/ml/datasets/Adult}}.
    The task is to predict whether the salary of a American citizen exceeds 50 thousand dollars per year based on census data. It contains more than 32000 instances.
    Each data instance is described by 14 features out of which 3 features are considered as sensitive, namely ``MaritalStatus'', ``Race'', ``Sex''.
  
  \textbf{LSAC.}
    This dataset contains information about 26551 law students\footnote{\url{http://www.seaphe.org/databases.php}}.
    The task is to classify whether law students pass the bar exam based on information from the study of the Law School Admission Council that was collected between 1991 and 1997.
    Each student is characterized by 11 features out of which 3 features, namely ``race'', ``sex'', ``family\_income'', are considered as sensitive.

We split each dataset into 70\% for the training set and 30\% for testing.
We then applied the Synthetic Minority Oversampling Technique (SMOTE\footnote{\url{https://machinelearningmastery.com/threshold-moving-for-imbalanced\\-classification/}}) over training data as class labels are highly imbalanced.
In particular, SMOTE generates samples to balance the class label distribution in order to overcome the imbalanced distribution problem.


\subsection{Tabular data: Selection of instances to assess fairness} \label{sec:exp_sampling_size}


We experimented different relative sizes for sampling instances.
The idea is to verify whether at least one sensitive feature appears in the list of the top-10 most important features.
For that, we used the following the relative sizes: 0.1\%, 0.5\%,  1\%, 5\%, and 10\%.
The results obtained throughout the performance of the same experiment 50 times are shown in Table \ref{tb:size_sampling}.

For the German dataset, we did not evaluate models using 0.1\% of the instances, as it takes only one instance.  On the other hand, for the Adult dataset, the amount of data instances from sampling 10\% of instances is large and unfeasible to produce many explanations (more than 3K) in time. We thus add an hyphen in Table \ref{tb:size_sampling} to indicate these particular cases.

\begin{table}[htpb]
\caption{%
Number of times pre-trained models are deemed unfair.
Experiments were performed by varying the number of instances selected to obtain global explanations, the sampling strategy, and the explanation method.}
\centering
\setlength{\tabcolsep}{5.1pt}
\begin{tabular}{|r|r|lllll|}
\hline
\multirow{2}{*}{\textbf{Dataset}} & \multirow{2}{*}{\textbf{Selection}} & \multicolumn{5}{c|}{\textbf{Sample size}} \\
  &   & 0.1\% & 0.5\% & 1\% & 5\% & 10\%  \\
\hline
\multirow{4}{*}{German} & \textit{LIME+RS} &   -   & 38 & 38 & 47 & 50 \\
                        & \textit{LIME+SP} &   -   & 47 & 47 & 50 & 50 \\
                        & \textit{SHAP+RS} &   -   & 50 & 50 & 50 & 50 \\
                        & \textit{SHAP+SP} &   -   & 50 & 50 & 50 & 50 \\
\hline
\multirow{4}{*}{Adult}  & \textit{LIME+RS} & 46 & 48 & 49 & 45 & - \\
                        & \textit{LIME+SP} & 49 & 49 & 49 & 49 & - \\
                        & \textit{SHAP+RS} & 49 & 47 & 46 & 47 & - \\
                        & \textit{SHAP+SP} & 47 & 50 & 49 & 49 & -  \\
\hline
\multirow{4}{*}{LSAC}   & \textit{LIME+RS} & 50 & 50 & 50 &  50 &  50 \\
                        & \textit{LIME+SP} & 50 & 50 & 50 &  50 &  50 \\
                        & \textit{SHAP+RS} & 50 & 50 & 50 &  50 &  50 \\
                        & \textit{SHAP+SP} & 50 & 50 & 50 &  50 &  50 \\
\hline
\end{tabular}
\label{tb:size_sampling}
\end{table}

To illustrate the impact of the sample size and the sampling method in the assessment of fairness, we observed the rank of sensitive features in the list of 10 most important features obtained by each of the explainers. These results are presented in Tables~\ref{tab:tabular-L-german},~\ref{tab:tabular-L-adult}, and~\ref{tab:tabular-L-lsac} for the datasets German, Adult, and LSAC, respectively.


\begin{table*}[htpb]
\centering
\setlength{\tabcolsep}{3pt}
\caption{Rank of sensitive features in $L$ (and their average contribution): Effects of the variation of the sample size on the German dataset.}
\label{tab:tabular-L-german}
\begin{tabular}{|c|r|r|rl|rl|rl|rl|rl|}
\hline
\multirow{2}{*}{}        & \multicolumn{1}{r|}{Sensitive} & \multicolumn{1}{c|}{\multirow{2}{*}{Selection}} & \multicolumn{10}{c|}{Sample size} \\
                         & \multicolumn{1}{r|}{feature}   & \multicolumn{1}{c|}{}                           & \multicolumn{2}{c}{0.1\%} & \multicolumn{2}{c}{0.5\%} & \multicolumn{2}{c}{1\%} & \multicolumn{2}{c}{5\%} & \multicolumn{2}{c|}{10\%} \\
\hline
\multirow{12}{*}{\rotcell[c]{German}} & \multirow{4}{*}{sex}  & \textit{LIME+RS}                            & -           & (-)           & 5             & (-0.015)    & 4             & (-0.117)   & 2             & (-1.630)   & {9}   & (-2.568)   \\
                         &                               & \textit{LIME+SP}                                 & -           & (-)          & 6             & (-0.147)    & 3             & (-0.397)   & 6             & (-1.610)   & 4            & (-3.063)   \\
                         &                               & \textit{SHAP+RS}                                      & -           & (-)           & {1}    & (-0.345)    & {1}    & (-0.345)   & {0}    & (-3.135)   & {0}   & (-6.836)   \\
                         &                               & \textit{SHAP+SP}                                        & -           & (-)           & {0}    & (0.360)     & {0}    & (0.617)    & {0}    & (3.123)    & {0}   & (6.853)    \\
\cline{2-13}
                         & \multirow{4}{*}{telephone}        & \textit{LIME+RS}                                      & -           & (-)           & 3             & (0.003)     & {8}    & (0.095)    & 6             & (1.509)    & {9}   & (2.891)    \\
                         &                               & \textit{LIME+SP}                                        & -           & (-)           & 3             & (0.023)     & 5             & (0.256)    & 6             & (1.431)    & 7            & (2.261)    \\
                         &                               & \textit{SHAP+RS}                                      & -           & (-)           & 7             & (0.121)     & 7             & (0.121)    & 5             & (1.245)    & 5            & (2.753)    \\
                         &                               & \textit{SHAP+SP}                                        & -           & (-)           & 2             & (-0.156)    & 4             & (-0.328)   & 5             & (-1.180)   & 4            & (-2.428)   \\
\cline{2-13}
                         & \multirow{4}{*}{foreign}      & \textit{LIME+RS}                                      & -           & (-)           & 2             & (0.041)     & {9}    & (0.076)    & {8}    & (1.078)    & 6            & (3.294)    \\
                         &                               & \textit{LIME+SP}                                        & -           & (-)           & {1}    & (0.376)     & {0}    & (0.721)    & {0}    & (3.289)    & {0}   & (7.103)    \\
                         &                               & \textit{SHAP+RS}                                      & -           & (-)           & {9}    & (0.122)     & {9}    & (0.122)    & {9}    & (0.490)    & *            & *        \\
                         &                               & \textit{SHAP+SP}                                       & -           & (-)           & 7             & (-0.100)    & {9}    & (-0.114)   & {9}    & (-0.444)   & {9}   & (-1.011)  \\
\hline
\end{tabular}
\end{table*}

\begin{table*}[htpb]
\centering
\setlength{\tabcolsep}{3pt}
\caption{Rank of sensitive features in $L$ (and their average contribution): Effects of the variation of the sample size on the Adult dataset.}
\label{tab:tabular-L-adult}
\begin{tabular}{|c|r|r|rl|rl|rl|rl|rl|}
\hline
\multirow{2}{*}{}        & \multicolumn{1}{r|}{Sensitive} & \multicolumn{1}{c|}{\multirow{2}{*}{Selection}} & \multicolumn{10}{c|}{Sample size} \\
                         & \multicolumn{1}{r|}{feature}   & \multicolumn{1}{c|}{}                           & \multicolumn{2}{c}{0.1\%} & \multicolumn{2}{c}{0.5\%} & \multicolumn{2}{c}{1\%} & \multicolumn{2}{c}{5\%} & \multicolumn{2}{c|}{10\%} \\
\hline
\multirow{12}{*}{\rotcell[c]{Adult}} & \multirow{4}{*}{marital} & \textit{LIME+RS}   & {9} & (-0.122) & {9} & (-0.012) & {8} & (0.138)  & {9}      &  (2.146)           & -           & (-)       \\
                        &                          & \textit{LIME+SP}   & 6          & (0.042)  & {9} & (0.367)  & {8} & (0.244)  & {8} & (2.746)  & -           & (-)       \\
                        &                          & \textit{SHAP+RS}   & 7          & (-0.003) & {9} & (-0.239) & {8} & (-0.246) & 7          & (-0.678) & -           & (-)       \\
                        &                          & \textit{SHAP+SP}   & {8} & (0.030)  & {9} & (0.093)  & 7          & (0.296)  & {9} & (1.731)  & -           & (-)       \\
\cline{2-13}
                        & \multirow{4}{*}{race}    & \textit{LIME+RS}   & {9} & (0.012)  & 6          & (-0.088) & 6          & (-0.253) & 7           & (-0.093)       & -           & (-)       \\
                        &                          & \textit{LIME+SP}   & {9} & (-0.009) & {8} & (-0.018) & 7          & (0.002)  & 7          & (-0.002) & -           & (-)       \\
                        &                          & \textit{SHAP+RS}   & {9} & (0.000)  & 6          & (0.000)  & {8} & (0.000)  & 6          & (0.000)  & -           & (-)       \\
                        &                          & \textit{SHAP+SP}   & 5          & (0.000)  & 7          & (0.000)  & 7          & (0.001)  & 7          & (0.000)  & -           & (-)       \\
\cline{2-13}
                        & \multirow{4}{*}{sex}     & \textit{LIME+RS}   & 4          & (-0.022) & 7          & (0.305)  & {8} & (0.617)  & {8}           & (4.902)       & -           & (-)       \\
                        &                          & \textit{LIME+SP}   & 6          & (0.153)  & 7          & (0.823)  & 7          & (0.484)  & 6          & (6.925)  & -           & (-)       \\
                        &                          & \textit{SHAP+RS}   & 7          & (0.050)  & 6          & (0.513)  & 6          & (1.138)  & 6          & (2.787)  & -           & (-)       \\
                        &                          & \textit{SHAP+SP}   & 7          & (-0.113) & 6          & (-0.203) & 6          & (-0.819) & 6          & (-3.387) & -           & (-)       \\
\hline
\end{tabular}
\end{table*}

\begin{table*}[htpb]
\centering
\setlength{\tabcolsep}{3pt}
\caption{Rank of sensitive features in $L$ (and their average contribution): Effects of the variation of the sample size on the LSAC dataset.}
\label{tab:tabular-L-lsac}
\begin{tabular}{|c|r|r|rl|rl|rl|rl|rl|}
\hline
\multirow{2}{*}{}        & \multicolumn{1}{r|}{Sensitive} & \multicolumn{1}{c|}{\multirow{2}{*}{Selection}} & \multicolumn{10}{c|}{Sample size} \\
                         & \multicolumn{1}{r|}{feature}   & \multicolumn{1}{c|}{}                           & \multicolumn{2}{c}{0.1\%} & \multicolumn{2}{c}{0.5\%} & \multicolumn{2}{c}{1\%} & \multicolumn{2}{c}{5\%} & \multicolumn{2}{c|}{10\%} \\
\hline
\multirow{12}{*}{\rotcell[c]{LSAC}}                        &                                                      & \textit{LIME+RS} & {9}                & (0.045)                & {9}                & (-0.962)               & {9}               & (-1.808)               & {9}   & (-6.754)    & {9}   & (-15.629)  \\
                        &                                                      & \textit{LIME+SP}   & {9}                & (-0.143)               & {9}                & (-0.948)               & {9}               & (-2.004)               & {9}   & (-7.074)    & {9}   & (-15.194)  \\
                        &                                                      & \textit{SHAP+RS} & 3                         & (-0.445)               & 2                         & (-1.986)               & {1}               & (-4.071)               & {1}   & (-18.311)   & {1}   & (-31.343)  \\
                        & \multirow{-4}{*}{sex}                                & \textit{SHAP+SP}   & 2                         & (0.477)                & {1}                & (1.827)                & {1}               & (4.943)                & {1}   & (14.738)    & {1}   & (46.189)   \\
\cline{2-13}
                        &                                                      & \textit{LIME+RS} & 6                         & (0.735)                & 3                         & (6.309)                & 3                        & (13.561)               & 6            & (67.047)    & 6            & (129.130)  \\
                        &                                                      & \textit{LIME+SP}   & 5                         & (1.222)                & 3                         & (6.827)                & 6                        & (12.202)               & 3            & (65.198)    & 6            & (132.644)  \\
                        &                                                      & \textit{SHAP+RS} & 4                         & (0.346)                & 3                         & (1.633)                & 2                        & (2.886)                & 3            & (14.858)    & 3            & (29.474)   \\
                        & \multirow{-4}{*}{race}                               & \textit{SHAP+SP}   & 7                         & (-0.275)               & 2                         & (-1.385)               & 3                        & (-2.724)               & 3            & (-15.227)   & 3            & (-27.966)  \\
\cline{2-13}
                        &                                                      & \textit{LIME+RS} & {8}                & (-0.289)               & {9}                & (-1.901)               & {9}               & (-2.342)               & {9}   & (-16.258)   & 7            & (-45.129)  \\
                        &                                                      & \textit{LIME+SP}   & {9}                & (-0.454)               & {9}                & (-2.250)               & 7                        & (-4.351)               & 7            & (-25.473)   & 7            & (-44.837)  \\
                        &                                                      & \textit{SHAP+RS} & {8}                & (0.119)                & {8}                & (0.487)                & {9}               & (0.812)                & 4            & (3.913)     & 3            & (11.642)   \\
 & \multirow{-4}{*}{f\_income}                          & \textit{SHAP+SP}   & {8}                & (-0.029)               & 6                         & (-0.568)               & 7                        & (-0.989)               & 6            & (-4.576)    & 6            & (-6.047)  \\
\hline
\end{tabular}
\end{table*}

We considered the following classifiers, namely Logistic Regression (LR), Bagging (BAG), AdaBoost (ADA), and Random Forest (RF), to perform the experiments.
We used the implementation available on Scikit-learn~\cite{scikit-learn} with the default parameters.
Based on these results we thus set the size as 5\% for the German dataset (50 instances), 0.5\%  for the Adult dataset (162 instances), and 0.1\% for the LSAC dataset (93 instances).
These results seem to indicate that LIME is more stable than SHAP independently of the sampling method and sample size.


\subsection{Tabular data: Automating the choice of $k$} \label{sec:exp_auto_k}

We evaluate the algorithm \FK that was conceived to automatically selects the list of features which is taken into account by \F for assessing fairness.
\FK requires a positive value for the parameter $\alpha$.
Here, we performed experiments by varying the value of $\alpha$ from 0.5 to 3.
In these experiments we analyze the impact of this parameter on the output of \FK. 
In particular, we are interested in the effects of different values of $\alpha$ w.r.t. the size of $L$ (i.e., the number of features kept by \FK) and record the number of times a model is deemed unfair. 
We discarded values of $\alpha < 0.5$ otherwise \FK considers all features as important.  

\begin{table*}[]
\centering
\setlength{\tabcolsep}{2.6pt}
\caption{Average value of $k$ for LR and BAG classifiers (number of times models are deemed unfair): Effects of the variation of $\alpha$ over the number of features kept by \FK.}
\label{tb:k_LR_BAG}
\begin{tabular}{|r|r|rl|rl|rl|rl|rl|rl||rl|rl|rl|rl|rl|rl|}
\hline
 \multirow{3}{*}{\textbf{\rotcell[c]{Data.}}}            & \multirow{3}{*}{\textbf{Selection}} & \multicolumn{12}{c||}{LR} & \multicolumn{12}{c|}{BAG} \\
 \cline{3-26}
 & & \multicolumn{12}{c||}{$\alpha$}  & \multicolumn{12}{c|}{$\alpha$}    \\
                  &  & \multicolumn{2}{c}{0.5}   & \multicolumn{2}{c}{1}     & \multicolumn{2}{c}{1.5}   & \multicolumn{2}{c}{2}     & \multicolumn{2}{c}{2.5}   & \multicolumn{2}{c||}{3}& \multicolumn{2}{c}{0.5}   & \multicolumn{2}{c}{1}     & \multicolumn{2}{c}{1.5}   & \multicolumn{2}{c}{2}     & \multicolumn{2}{c}{2.5}   & \multicolumn{2}{c|}{3}    \\
\hline
 \multirow{4}{*}{\rotcell[c]{German}} & \textit{LIME+RS} & {9.9} & ({47}) & {8.7} & (42) & {6.8} & (31) & 5.5 & (23) & 3.5 & (13) & 2.6 & (8) & 10.0 & (33) & 9.58 & (31) & 8.18 & (24) & 6.38 & (18) & 4.26 & (8) & 2.82 & (2) \\
                         & \textit{LIME+SP}   & {9.7} & ({50})  & {8.2} & ({49})  & {6.1} & (43)  & {4.5} & (39)  & 2.1 & (29)  & 1.3 & (24) & {10.0} & (41) & {10.0} & (41) & {9.74} & (41) & 9.08 & (40) & 7.04 & (30) & 4.86 & (21) \\
                         & \textit{SHAP+RS}   & {10.0} & ({50}) & {9.9} & ({50})  & {9.5} & ({50})  & {7.4} & ({47})  & {5.7} & ({45})  & {3.8} & (43)  & 9.92 & (37) & 9.80 & (37) & 8.78 & (33) & 7.16 & (26) & 5.86 & (22) & 4.86 & (19) \\
                         & \textit{SHAP+SP}   & {10.0}  & ({50}) & {9.8}  & ({50}) & {8.9}  & ({50}) & {7.0}  & ({49}) & {5.4}  & ({45}) & {4.0}  & (43) & {10.0} & ({45}) & {9.78} & ({45}) & {9.12} & (43) & 7.72 & (36) & 6.58 & (31) & 5.50 & (26) \\
\hline
 \multirow{4}{*}{\rotcell[c]{Adult}}  & \textit{LIME+RS}   & {10.0} & ({46}) & {10.0} & ({46}) & {10.0} & ({46}) & {9.9} & ({46}) & {9.9}  & (44) & {9.8}  & (44) & {10.0} & ({49}) & {10.0} & ({49}) & {10.0} & ({49}) & {9.58} & ({49}) & {8.32} & ({45}) & 6.82 & (40) \\
                         & \textit{LIME+SP}   & {10.0} & ({49}) & {10.0} & ({49}) & {10.0} & ({49}) & {10.0} & ({49})  & {9.9} & ({49}) & {9.8} & ({49}) & {10.0} & (50) & {10.0} & (50) & {10.0} & (50) & {9.64} & (50) & {8.52} & (50) & {7.32} & (50) \\
                         & \textit{SHAP+RS}   & {10.0} & ({48}) & {10.0} & ({48})  & {10.0} & ({48})  & {10.0} & ({48})  & {9.8} & ({48})  & {9.7} & ({45})  & {10.0} & ({49}) & {10.0} & ({49}) & {10.0} & ({49}) & {9.62} & ({49}) & {8.30} & ({45}) & 7.08 & (40) \\
                         & \textit{SHAP+SP}   & {10.0} & ({50}) & {9.9} & ({50}) & {9.8} & ({50}) & {9.7} & ({50}) & {9.6} & ({49}) & {9.3} & ({49}) & {10.0} & ({50}) & {10.0} & ({50}) & {8.84} & ({50}) & 5.84 & (39) & 2.32 & (11) & 1.32 & (2) \\
\hline
\multirow{4}{*}{\rotcell[c]{LSAC}}   & \textit{LIME+RS}   & {8.6} & ({47})  & 6.4 & (28)  & 4.0 & (10)  & 2.5 & (6)  & 1.8 & (1)  & 1.3 & (0) & {7.74} & ({50})  & {5.64} & ({49}) & {4.04} & (43) & {2.68} & (32) & 1.98 & (24) & 1.56 & (21) \\
                         & \textit{LIME+SP}   & 7.7 & (35) & 5.1 & (22) & 3.1 & (14) & 1.7 & (4) & 1.1 & (0) & 1.0 & (0) & {9.72} & ({50})  & {8.76} & ({50}) & {7.86} & ({50}) & {7.00} & ({50}) & {5.96} & ({49}) & {4.84} & (44) \\
                         & \textit{SHAP+RS}   & {8.5} & ({48}) & {6.6} & (42) & 4.5 & (30) & 3.2 & (22) & 2.5 & (18) & 2.0 & (12) & {8.34} & ({50}) & {5.98} & (32) & 4.46 & (24) & 3.04 & (17) & 2.56 & (14) & 2.08 & (11) \\
                         & \textit{SHAP+SP}   & {8.3} & ({50}) & {6.1} & ({45}) & {4.4} & (36) & 3.2 & (26) & 2.3 & (19) & 1.7 & (14) & {8.70} & ({49})  & {6.88}  & (44) & {5.10} & (33) & 3.76 & (26) & 2.82 & (14) & 2.22 & (11)\\
\hline
\end{tabular}
\end{table*}

\begin{table*}[]
\centering
\setlength{\tabcolsep}{2.6pt}
\caption{Average value of $k$ for RF and ADA classifiers (number of times models are deemed unfair): Effects of the variation of $\alpha$ over the number of features kept by \FK.}
\label{tb:k_RF_ADA}
\begin{tabular}{|r|r|rl|rl|rl|rl|rl|rl||rl|rl|rl|rl|rl|rl|}
\hline
 \multirow{3}{*}{\textbf{\rotcell[c]{Data.}}}            & \multirow{3}{*}{\textbf{Selection}} & \multicolumn{12}{c||}{RF} & \multicolumn{12}{c|}{ADA} \\
 \cline{3-26}
 & & \multicolumn{12}{c||}{$\alpha$}  & \multicolumn{12}{c|}{$\alpha$}    \\
                  &  & \multicolumn{2}{c}{0.5}   & \multicolumn{2}{c}{1}     & \multicolumn{2}{c}{1.5}   & \multicolumn{2}{c}{2}     & \multicolumn{2}{c}{2.5}   & \multicolumn{2}{c||}{3}& \multicolumn{2}{c}{0.5}   & \multicolumn{2}{c}{1}     & \multicolumn{2}{c}{1.5}   & \multicolumn{2}{c}{2}     & \multicolumn{2}{c}{2.5}   & \multicolumn{2}{c|}{3}    \\
\hline
 \multirow{4}{*}{\rotcell[c]{German}} & \textit{LIME+RS}   & {9.90} & ({49}) & {8.30} & (43) & 5.18 & (25) & 2.72 & (8) & 1.54& (2) & 1.18 & (2) & {10} & ({48})  & {10} & ({48})  & {9.76} & ({48}) & {9.4} & ({46})  & {9.2} & ({45})   & {9.04} & (44)\\
                         & \textit{LIME+SP}   & {10.0} & ({50}) & {9.98} & ({50}) & {9.74} & ({50}) & {8.54} & ({50}) & {6.96} & ({47}) & {5.46} & ({46}) & {10.0} & ({50}) & {10.0} & ({50}) & {10.0} & ({50}) & {10.0} & ({50}) & {10.0} & ({50}) & {10.0} & ({50})\\
                         & \textit{SHAP+RS}   & {9.92} & ({50})  & {8.98} & ({48}) & {6.46} & (39) & 4.52 & (29) & 2.74 & (23) & 2.04 & (18) & {10.0} & ({48}) & {8.78} & (43) & 6.44 & (31) & 4.28 & (23) & 3.24 & (15) & 2.90 & (13) \\
                         & \textit{SHAP+SP}   & {9.86} & ({50}) & {8.70} & ({47}) & 5.64 & (33) & 3.68 & (27) & 2.28 & (22) & 1.42 & (17) & {9.98} & ({47}) & {8.68} & (42) & 6.78 & (34) & 5.82 & (30) & 4.92 & (24) & 4.04 & (20)\\
\hline
 \multirow{4}{*}{\rotcell[c]{Adult}}  & \textit{LIME+RS}   & {9.76} & ({50}) & {8.38} & ({50}) & {7.00} & ({48}) & {5.48} & (38) & 4.44 & (18) & 3.64 & (6) & {10.0} & ({50}) & {10.0} & ({50}) & {8.22} & ({49}) & {5.54} & (35) & {3.40} & (13) & {2.20} & (2) \\
                         & \textit{LIME+SP}   & {9.30} & ({50}) & {7.80} & ({50}) & {6.76} & ({50}) & {5.80} & ({48}) & 5.00 & (40) & 4.32 & (27) & {10.0} & ({50}) & {9.90} & ({50}) & {7.98} & ({49}) & {5.74} & (43) & {3.84} & (19) & {2.48} & (4) \\
                         & \textit{SHAP+RS}   & {10.0} & ({50}) & {9.96} & ({50}) & {9.30} & ({50}) & {8.12} & ({50}) & {6.48} & ({46}) & {5.14} & (41) & {10.0} & ({50}) & {9.02} & ({50}) & {6.62} & ({49}) & {4.76} & ({48}) & {3.28} & ({47}) & {2.38} & ({47}) \\
                         & \textit{SHAP+SP}   & {10.0} & ({50})  & {9.98} & ({50})  & {9.38} & ({50}) & {8.02} & ({48}) & {6.26} & ({47}) & {4.94} & (38) & {10.0} & ({50}) & {9.16} & ({49}) & {7.22} & ({49}) & {5.36} & ({49}) & {4.06} & ({47}) & {2.82} & ({47}) \\
\hline
 \multirow{4}{*}{\rotcell[c]{LSAC}}   & \textit{LIME+RS}   & {6.46} & ({49}) & {4.04} & (38) & 2.02 & (16) & 1.46 & (10) & 1.10 & (6) & 1.02 & (6) & {6.98} & (44) & {4.52} & (32) & 3.02 & (20) & 1.92 & (11) & 1.28 & (5) & 1.12 & (4) \\
                         & \textit{LIME+SP}   & {8.92} & ({50})  & {7.30} & ({49}) & {5.38} & ({48}) & {3.66} & ({48}) & 2.66 & ({48}) & 1.96 & ({48}) & {7.08} & ({46}) & {5.06} & (34) & 3.44 & (22) & 2.22 & (11) & 1.78 & (9) & 1.28 & (4)  \\
                         & \textit{SHAP+RS}   & {7.80} & ({47}) & {5.80} & (40) & 3.88 & (21) & 2.48 & (13) & 1.76 & (7) & 1.40 & (3) & {8.68} & ({50}) & {6.52} & (42) & 4.78 & (30) & 3.34 & (21) & 2.08 & (8) & 1.52 & (6)  \\
                         & \textit{SHAP+SP}   & {8.18} & ({49}) & {5.78} & (37) & 4.04 & (23) & 2.86 & (13) & 2.10 & (8) & 1.70 & (4) & {8.84} & ({50}) & {7.08} & ({47}) & {5.42} & (39) & 3.84 & (26) & 2.68 & (12) & 1.90 & (7)  \\
\hline
\end{tabular}
\end{table*}

High values of $\alpha$ lead to a smaller size of $L$, i.e., too many features are removed by \FK. It is harder to find one sensitive feature in a very short list of important features than in a longer list.
As a consequence, high values of $\alpha$ also lead to a low number of times in which a model is deemed unfair.
We can observe this behavior on Tables~\ref{tb:k_LR_BAG} and~\ref{tb:k_RF_ADA}.
These tables contain the number of times that the models were deemed unfair throughout 50 repetitions of the same experiment.

A similar behavior can be observed w.r.t. the average value of $k$.
High values of $\alpha$ lead to low values of $k$ (on average).
We notice this behavior in Tables~\ref{tb:k_LR_BAG} and~\ref{tb:k_RF_ADA}. These tables contain the average value of $k$ found by \FK.
We highlighted the experiments where the average number of sensitive features found in the list obtained by \F was greater than or equal to 1.
Again, \F looks for at least one sensitive feature in the list of the most important features.
Then, an average smaller than 1 indicates that in most experiments the models were deemed fair.
We also notice that the highlighted values are those in which the average value of $k$ are high (see Tables~\ref{tb:k_LR_BAG} and~\ref{tb:k_RF_ADA}).

We can observe that \FK discovers on average the same value of $k$ for each dataset.
For instance, for the German dataset, an average $k$ around 10 was found by \FK with $\alpha=0.5$, and incidentally the same value was used in previous empirical studies~\cite{bhargava,alves}.
For the LSAC dataset, an average $k$ around 8 was found using \FK with $\alpha=0.5$, while an average $k$ around 6 was returned by \FK using $\alpha=1$.

To sum up the analysis of \FK, we can also observe that low values of $\alpha$ provide suitable configurations for users that do not have any clue of the useful values of $k$.
More precisely, using $0.5\leq \alpha \leq 1$ allows \FK to automatically find a suitable value of $k$.
In addition, since LIME and SHAP have stability issues \cite{slack2020fooling} that lead to different lists of feature importance (for a single type of classifier trained on the same dataset), it may be interesting 
to considering the threshold $\alpha$ to learn the parameter $k$. Indeed,
one single value of $\alpha$ allows \FK to find a suitable value of $k$ from distinct lists of feature importance.  
This provides an answer to question Q1 introduced at the beginning of this paper.

%




\subsection{Textual datasets} \label{sec:exp_text_data}

We also experimented \F with simple textual classifiers.
In particular, we implemented a model used in an experiment carried out by Davidson et al. \cite{davidson2019racial}, whose goal is to classify tweets as \textit{hate speech} or not.
We focus on the \textit{hate speech} dataset \cite{hateoffensive}, which more precisely labels a tweet as \textit{offensive}, \textit{hate speech}, or \textit{neither}.
To stay within a two-classes problem, we merge \textit{offensive language} and \textit{hate speech} classes, to finally deal with two classes, namely \textit{non-offensive} or \textit{offensive}.

However, the resulting dataset appears to be quite unbalanced, including 4163 non-offensive instances and 20620 offensive ones.
Thus we randomly select only 4163 instances from the offensive group.
Moreover, we follow the same representation and pre-processing steps as in \cite{davidson2019racial}. We stem each word in order to reduce and to gather
 similar words, e.g., \textit{``vehicle''} and \textit{``vehicles''} are both transformed into \textit{``vehicl''}.
The classifier is based on a TF-IDF vectorizer followed by a logistic regression.
In addition, we ran experiments with a bagging ensemble, a random forest, and AdaBoost, instead of the logistic regression.

Then we apply the first step of \F to build a global explanation.
This is performed with the textual adaptation of LIME and SHAP.
We compare  both random sampling (RS) and submodular pick (SP), for these two explainers.
Unlike the previous tabular classifiers, the model deals with more than 2000 features, i.e., as many as the vocabulary size.
For example, Table~\ref{tab:global-lr} shows the first 10 most important words obtained after 50 experiments run with a logistic regression and LIME explanations, respectively with random sampling and submodular pick.

\begin{table}[!ht]
    \centering
    \caption{Mean contribution of a logistic regression, LIME as explainer, after 50 experiments}
    \begin{tabular}{|c|c|c|}
        \hline
        \multicolumn{3}{|c|}{\textit{Random sampling}} \\
        \hline
        \textbf{rank} & \textbf{word} & \textbf{contrib.} \\
        \hline
        1 & faggot & 0.632 \\
        2 & fag & 0.625 \\
        3 & bitch & 0.622 \\
        4 & niggah & 0.620 \\
        5 & cunt & 0.613 \\
        6 & pussi & 0.608 \\
        7 & nigger & 0.597 \\
        8 & hoe & 0.580 \\
        9 & nigguh & 0.522 \\
        10 & dyke & 0.473 \\
        \hline
    \end{tabular}
    \hspace{10pt}
    \begin{tabular}{|c|c|c|}
        \hline
        \multicolumn{3}{|c|}{\textit{Submodular pick}} \\
        \hline
        \textbf{rank} & \textbf{word} & \textbf{contrib.} \\
        \hline
        1 & niggah & 0.596 \\
        2 & cunt & 0.579 \\
        3 & bitch & 0.574 \\
        4 & fag & 0.573 \\
        5 & faggot & 0.573 \\
        6 & nigger & 0.572 \\
        7 & pussi & 0.547 \\
        8 & hoe & 0.524 \\
        9 & nigguh & 0.505 \\
        10 & retard & 0.414 \\
        \hline
    \end{tabular}
    \label{tab:global-lr}
\end{table}

At first glance, we can observe that important words for hate speech classification are insults and swear words, which makes sense.
However, it should be noticed that such a classification may also be related to a particular context.
For example, the use of words such as \textit{``n*gger''} in a conversation may also be related to a familiar interaction between two very close friends, and thus not at all indicate an \textit{offensive} discussion.



\begin{figure}[!ht]
    \centering
    \includegraphics[width=0.85\linewidth]{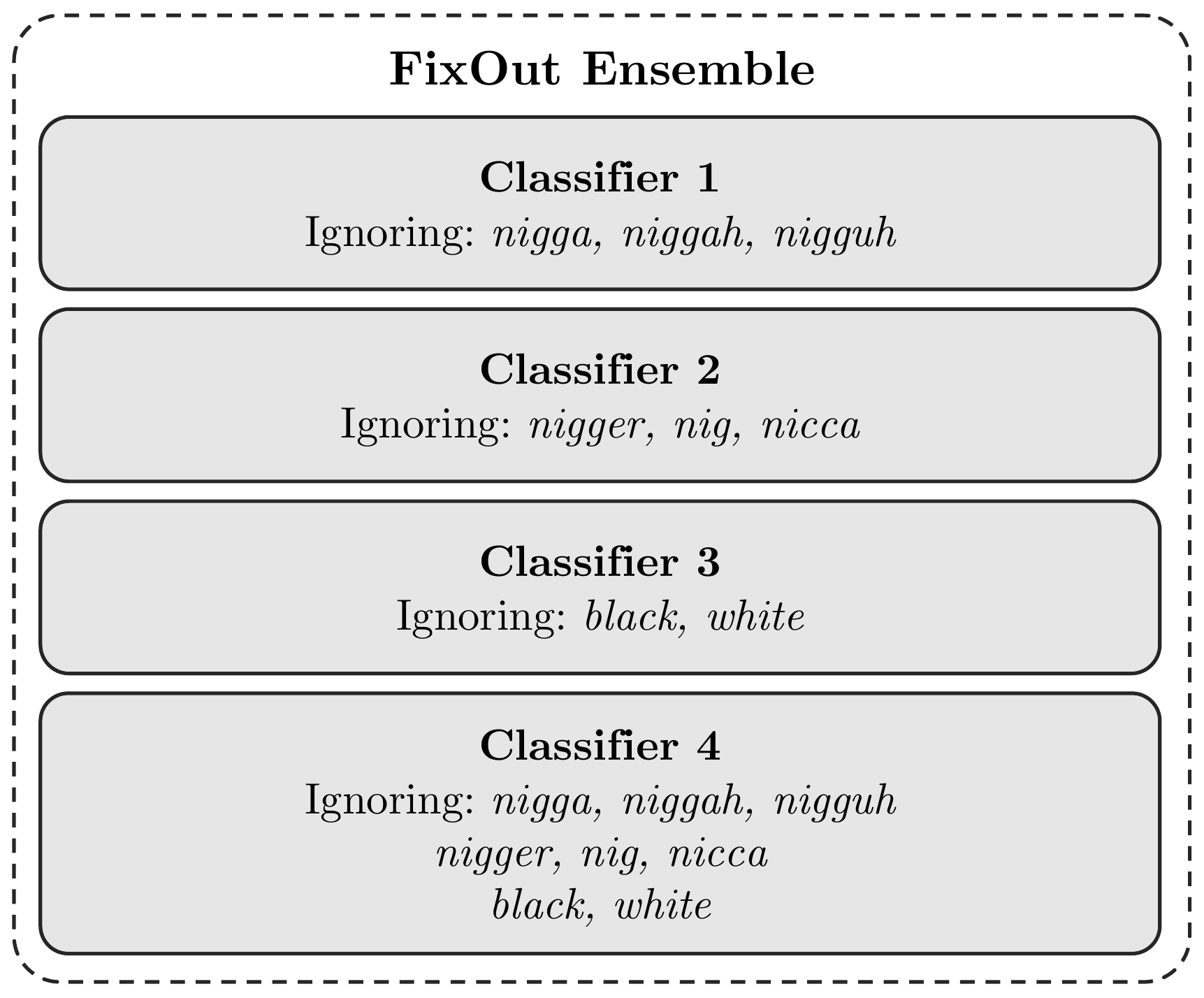}
    \caption{Illustration of textual classifiers used in the ensemble.}
    \label{fig:text-ensemble}
\end{figure}

This shows that the definition of a \textit{sensitive word} is not straightforward in such a case, and that we must find words that are responsible for a bias.
Then, in order to complete this experiment, we manually select sensitive features.
Accordingly, the objective of the next subsection is to demonstrate that \F is able to decrease the contribution of pre-selected words thanks to a \F ensemble and by grouping words.
For example, considering the groups of words presented in Figure~\ref{fig:text-ensemble},  the goal of this ensemble is to reduce the contribution of these groups of words.


\subsection{Textual data: Process fairness assessment}
\label{sec:exp_text_proc_fairness}

For each experiment, we train 4 models, namely logistic regression, random forest, adaboost, and  bagging, plus 4 models corresponding to the related \F ensembles, which should be less dependent to sensitive words as defined above.
Next, we use LIME and SHAP explanations, with both random sampling and submodular pick.
For each explanation, we assess the contribution of all the features.
We then sort all the features w.r.t. their average contribution, select the 500 most important words, and then check their rank.
Words which are not in this list are therefore ranked beyond 500 and their contribution is marked as `` - '' to indicate that it is not significant.

Table~\ref{tab:textual} shows the results obtained with random sampling, including ranking difference.
LIME explanations systematically show a decrease in the contribution and the importance rank of selected words.
We even observe that the contribution is often nearly divided by 2, although the contribution of the whole vocabulary tends to decrease a bit in the ensemble explanation.
By contrast, SHAP explanations show an increase in the ranking of the word \textit{nigga}, although its contribution decreases.
The contribution of all other selected words is decreasing in general, even if this tendency seems to be more instable than in the case of LIME.
Table~\ref{tab:textual} also presents the results obtained with submodular pick, and shows results similar to random sampling with better results with logistic regression (+1.22 in average with LIME and +1.90 with SHAP).
These latter results seem to indicate that  \F with LIME explanations is more stable and consistently improves process fairness, i.e., always decrease the contribution of sensitive features.
However, \F with SHAP seems to provide greater improvements; see, e.g., ``nigguh'' and ``nig'' in the case of LR (diff. rank 22.81 and 54.56 with submodular pick and 21.73 and 40.67 with random sampling, resp.). It is noteworthy that for SHAP ``nigga'' always worsens.

\begin{table*}[!ht]
    \caption{Process fairness assessment on textual data, global explanations with submodular pick and random sampling, respectively.}
    \centering
    \setlength{\tabcolsep}{4pt}
    \begin{tabular}{|l|c||c|c||c|c||c|}
        \hline
        &  & \multicolumn{2}{c||}{{Original model}} & \multicolumn{2}{c||}{{\F Ensemble}} & Diff\\
        & \textbf{Word} & \textbf{Rank} & \textbf{Contrib.} & \textbf{Rank} & \textbf{Contrib.} & \textbf{Rank}\\
        \hline
        \multirow{8}{*}{\rotcell[c]{LR, LIME+SP}} & \textit{niggah} & 1 & 0.596 & 12 & 0.275 & 11\\
        & \textit{nigger} & 6 & 0.572 & 13 & 0.274 & 1.16\\
        & \textit{nigguh} & 9 & 0.505 & 14 & 0.269 & 0.55\\
        & \textit{nig} & 13 & 0.315 & 20 & 0.138 & 0.54\\
        & \textit{nicca} & 15 & 0.289 & 23 & 0.12 & 0.53\\
        & \textit{nigga} & 18 & 0.24 & 24 & 0.119 & 0.33\\
        & \textit{white} & 23 & 0.178 & 45 & 0.096 & 0.96\\
        & \textit{black} & 145 & 0.047 & 323 & 0.028 &1.23\\
        \hline
        \multirow{8}{*}{\rotcell[c]{RF, LIME+SP}} & \textit{niggah} & 7 & 0.517 & 12 & 0.257 &0.71\\
        & \textit{nigger} & 9 & 0.476 & 15 & 0.23 &0.67\\
        & \textit{nigguh} & 13 & 0.339 & 17 & 0.194&0.31 \\
        & \textit{nig} & 10 & 0.445 & 16 & 0.204 &0.60\\
        & \textit{nicca} & 16 & 0.265 & 20 & 0.121&0.25 \\
        & \textit{nigga} & 17 & 0.235 & 23 & 0.112 &0.35\\
        & \textit{white} & 23 & 0.127 & 34 & 0.07 &0.48\\
        & \textit{black} & $>$500 & $\sim 0$ & $>$500 & $\sim 0$ &0.00\\
        \hline
        \multirow{8}{*}{\rotcell[c]{ADA, LIME+SP}} & \textit{niggah} & 2 & 0.167 & 4 & 0.083 &1.00\\
        & \textit{nigger} & 7 & 0.052 & 10 & 0.026 &0.43\\
        & \textit{nigguh} & 5 & 0.144 & 6 & 0.073 &0.20\\
        & \textit{nig} & 18 & 0.014 & 24 & 0.006 &0.33\\
        & \textit{nicca} & 17 & 0.015 & 23 & 0.007 &0.35\\
        & \textit{nigga} & 4 & 0.166 & 5 & 0.083 &0.25\\
        & \textit{white} & 23 & 0.011 & 26 & 0.005 &0.13\\
        & \textit{black} & 113 & 0.0 & 196 & 0.0 &0.73\\
        \hline
        \multirow{8}{*}{\rotcell[c]{BAG, LIME+SP}} & \textit{niggah} & 5 & 0.645 & 15 & 0.293& 2.00\\
        & \textit{nigger} & 11 & 0.58 & 16 & 0.29 &0.45\\
        & \textit{nigguh} & 13 & 0.497 & 18 & 0.233 &0.38\\
        & \textit{nig} & 1 & 0.659 & 10 & 0.336 &9.00\\
        & \textit{nicca} & 15 & 0.387 & 19 & 0.195 &0.27\\
        & \textit{nigga} & 19 & 0.285 & 21 & 0.139 &0.11\\
        & \textit{white} & 21 & 0.183 & 24 & 0.083 &0.14\\
        & \textit{black} & 420 & 0.004 & $>$500 & $\sim 0$ & 0.19\\
        \hline
    \end{tabular}
    \hspace{10pt}
    \begin{tabular}{|l|c||c|c||c|c||c|}
        \hline
        &  & \multicolumn{2}{c||}{{Original model}} & \multicolumn{2}{c||}{{\F Ensemble}} & Diff\\
        & \textbf{Word} & \textbf{Rank} & \textbf{Contrib.} & \textbf{Rank} & \textbf{Contrib.} &\textbf{Rank}\\
        \hline
        \multirow{8}{*}{\rotcell[c]{LR, SHAP+SP}} & \textit{niggah} & 16 & 0.179 & 75 & 0.015 & 3.69\\
        & \textit{nigger} & 11 & 0.239 & 23 & 0.034 &1.09\\
        & \textit{nigguh} & 21 & 0.123 & $>$500 & $\sim 0$ &22.81\\
        & \textit{nig} & 9 & 0.249 & $>$500 & $\sim 0$ &54.56\\
        & \textit{nicca} & 22 & 0.12 & 46 & 0.02 &1,09\\
        & \textit{nigga} & 19 & 0.159 & 14 & 0.053 &-0.36\\
        & \textit{white} & 20 & 0.14 & 50 & 0.019 &1,50\\
        & \textit{black} & $>$500 & $\sim 0$ & $>$500 & $\sim 0$ &0.00\\
        \hline
        \multirow{8}{*}{\rotcell[c]{RF, SHAP+SP}} & \textit{niggah} & 14 & 0.176 & 23 & 0.03 &0.64\\
        & \textit{nigger} & 12 & 0.213 & 21 & 0.031 &0.75\\
        & \textit{nigguh} & 22 & 0.13 & 83 & 0.008 &2.77\\
        & \textit{nig} & 7 & 0.276 & 65 & 0.011 &8.29\\
        & \textit{nicca} & 21 & 0.138 & 39 & 0.018 &0.86\\
        & \textit{nigga} & 18 & 0.155 & 12 & 0.067 &-0.50\\
        & \textit{white} & 26 & 0.085 & 36 & 0.018 &0.38\\
        & \textit{black} & $>$500 & $\sim 0$ & $>$500 & $\sim 0$ &0.00\\
        \hline
        \multirow{8}{*}{\rotcell[c]{ADA, SHAP+SP}} & \textit{niggah} & 5 & 0.05 & 11 & 0.007&1.20 \\
        & \textit{nigger} & 9 & 0.018 & 13 & 0.005 &0.44\\
        & \textit{nigguh} & 6 & 0.039 & 25 & 0.002 &3.17\\
        & \textit{nig} & 18 & 0.009 & 86 & 0.001 &3.78\\
        & \textit{nicca} & 34 & 0.006 & 40 & 0.001&0.18 \\
        & \textit{nigga} & 4 & 0.07 & 3 & 0.031 &-0.33\\
        & \textit{white} & 32 & 0.007 & 30 & 0.002 &-0.07\\
        & \textit{black} & $>$500 & $\sim 0$ & $>$500 & $\sim 0$ &0.00\\
        \hline
        \multirow{8}{*}{\rotcell[c]{BAG, SHAP+SP}} & \textit{niggah} & 15 & 0.202 & 24 & 0.021 & 0.60\\
        & \textit{nigger} & 12 & 0.28 & 17 & 0.038 &0.42\\
        & \textit{nigguh} & 19 & 0.166 & 44 & 0.012 &1.32\\
        & \textit{nig} & 2 & 0.508 & 14 & 0.044 &6.00\\
        & \textit{nicca} & 14 & 0.209 & 21 & 0.025 &0.50\\
        & \textit{nigga} & 17 & 0.195 & 11 & 0.078 &-0.55\\
        & \textit{white} & 25 & 0.074 & 72 & 0.008 &1.88\\
        & \textit{black} & $>$500 & $\sim 0$ & $>$500 & $\sim 0$ &0.00\\
        \hline
    \end{tabular}
    \label{tab:textual}
\end{table*}
\noindent
\begin{table*}[!ht]
    \centering
    \setlength{\tabcolsep}{4pt}
    \begin{tabular}{|l|c||c|c||c|c||c|}
        \hline
        &  & \multicolumn{2}{c||}{{Original model}} & \multicolumn{2}{c||}{{\F Ensemble}} & Diff\\
        & \textbf{Word} & \textbf{Rank} & \textbf{Contrib.} & \textbf{Rank} & \textbf{Contrib.} & \textbf{Rank}\\
        \hline
        \multirow{8}{*}{\rotcell[c]{LR, LIME+RS }} & \textit{niggah} & 4 & 0.62 & 12 & 0.277 & 2.00\\
        & \textit{nigger} & 7 & 0.597 & 10 & 0.29 & 0.43\\
        & \textit{nigguh} & 9 & 0.522 & 14 & 0.264 & 0.56\\
        & \textit{nig} & 13 & 0.328 & 20 & 0.15 & 0.54\\
        & \textit{nicca} & 14 & 0.316 & 22 & 0.134 &0.57\\
        & \textit{nigga} & 18 & 0.237 & 27 & 0.113 &0.50\\
        & \textit{white} & 23 & 0.172 & 35 & 0.104 &0.52\\
        & \textit{black} & 131 & 0.05 & 316 & 0.028 &1.41\\
        \hline
        \multirow{8}{*}{\rotcell[c]{RF, LIME+RS }} & \textit{niggah} & 7 & 0.593 & 10 & 0.289 &0.43\\
        & \textit{nigger} & 9 & 0.497 & 14 & 0.236 &0.56\\
        & \textit{nigguh} & 13 & 0.358 & 17 & 0.198 &0.31\\
        & \textit{nig} & 10 & 0.46 & 16 & 0.222 &0.60\\
        & \textit{nicca} & 14 & 0.286 & 20 & 0.132 &0.43\\
        & \textit{nigga} & 18 & 0.23 & 23 & 0.112 &0.28\\
        & \textit{white} & 23 & 0.119 & 33 & 0.066 &0.43\\
        & \textit{black} & $>$500 & $\sim 0$ & $>$500 & $\sim 0$ &0.00\\
        \hline
        \multirow{8}{*}{\rotcell[c]{ADA, LIME+RS}} & \textit{niggah} & 3 & 0.166 & 5 & 0.082 &0.67\\
        & \textit{nigger} & 7 & 0.052 & 11 & 0.026 &0.57\\
        & \textit{nigguh} & 5 & 0.144 & 6 & 0.074 &0.20\\
        & \textit{nig} & 18 & 0.014 & 24 & 0.006 &0.33\\
        & \textit{nicca} & 17 & 0.015 & 23 & 0.007 &0.35\\
        & \textit{nigga} & 4 & 0.166 & 4 & 0.083 &0.00\\
        & \textit{white} & 22 & 0.011 & 26 & 0.005 &0.18\\
        & \textit{black} & 115 & 0.0 & 192 & 0.0 &0.67\\
        \hline
        \multirow{8}{*}{\rotcell[c]{BAG, LIME+RS}} & \textit{niggah} & 6 & 0.687 & 13 & 0.32 &1.17\\
        & \textit{nigger} & 10 & 0.629 & 15 & 0.306&0.50 \\
        & \textit{nigguh} & 13 & 0.535 & 17 & 0.248 &0.31\\
        & \textit{nig} & 2 & 0.7 & 10 & 0.353 &4.00\\
        & \textit{nicca} & 15 & 0.395 & 19 & 0.2 &0.27\\
        & \textit{nigga} & 19 & 0.284 & 21 & 0.139 &0.11\\
        & \textit{white} & 22 & 0.158 & 25 & 0.074 &0.14\\
        & \textit{black} & $>$500 & $\sim 0$ & $>$500 & $\sim 0$ &0,00\\
        \hline
    \end{tabular}
    \hspace{10pt}
    \begin{tabular}{|l|c||c|c||c|c||c|}
        \hline
        &  & \multicolumn{2}{c||}{{Original model}} & \multicolumn{2}{c||}{{\F Ensemble}} & Diff\\
        & \textbf{Word} & \textbf{Rank} & \textbf{Contrib.} & \textbf{Rank} & \textbf{Contrib.} &\textbf{Rank}\\
        \hline
        \multirow{8}{*}{\rotcell[c]{LR, SHAP+RS}} & \textit{niggah} & 16 & 0.194 & 68 & 0.016 &3.25\\
        & \textit{nigger} & 7 & 0.273 & 18 & 0.042 &1.57\\
        & \textit{nigguh} & 22 & 0.126 & $>$500 & $\sim 0$ &21.73\\
        & \textit{nig} & 12 & 0.223 & $>$500 & $\sim 0$ &40.67\\
        & \textit{nicca} & 21 & 0.138 & 37 & 0.024 &0.76\\
        & \textit{nigga} & 19 & 0.179 & 13 & 0.061 &-0.46\\
        & \textit{white} & 20 & 0.177 & 53 & 0.019 &1.65\\
        & \textit{black} & $>$500 & $\sim 0$ & $>$500 & $\sim 0$ &0.00\\
        \hline
        \multirow{8}{*}{\rotcell[c]{RF, SHAP+RS}} & \textit{niggah} & 14 & 0.188 & 22 & 0.03 &0.57\\
        & \textit{nigger} & 12 & 0.219 & 21 & 0.031 &0.75\\
        & \textit{nigguh} & 23 & 0.132 & 84 & 0.008 &2.65\\
        & \textit{nig} & 10 & 0.272 & 68 & 0.011 &5.80\\
        & \textit{nicca} & 21 & 0.154 & 36 & 0.021 &0.71\\
        & \textit{nigga} & 17 & 0.17 & 12 & 0.075 &-0.42\\
        & \textit{white} & 25 & 0.11 & 38 & 0.019 &0.52\\
        & \textit{black} & $>$500 & $\sim 0$ & $>$500 & $\sim 0$ &0.00\\
        \hline
        \multirow{8}{*}{\rotcell[c]{ADA, SHAP+RS}} & \textit{niggah} & 5 & 0.048 & 11 & 0.007 &1.20\\
        & \textit{nigger} & 9 & 0.019 & 13 & 0.005 &0.44\\
        & \textit{nigguh} & 6 & 0.039 & 24 & 0.002 &3.00\\
        & \textit{nig} & 20 & 0.008 & 80 & 0.001 &3.00\\
        & \textit{nicca} & 26 & 0.006 & 37 & 0.001 &0.42\\
        & \textit{nigga} & 4 & 0.072 & 3 & 0.031 &-0.33\\
        & \textit{white} & 22 & 0.008 & 28 & 0.002 &0.27\\
        & \textit{black} & $>$500 & $\sim 0$ & $>$500 & $\sim 0$ &0.00\\
        \hline
        \multirow{8}{*}{\rotcell[c]{BAG, SHAP+RS}} & \textit{niggah} & 17 & 0.203 & 26 & 0.02 &0.53\\
        & \textit{nigger} & 13 & 0.257 & 16 & 0.039 &0.23\\
        & \textit{nigguh} & 19 & 0.165 & 46 & 0.012 &1.42\\
        & \textit{nig} & 2 & 0.442 & 14 & 0.044 &6.00\\
        & \textit{nicca} & 14 & 0.217 & 21 & 0.026 &0.50\\
        & \textit{nigga} & 16 & 0.206 & 11 & 0.082 &-0.45\\
        & \textit{white} & 24 & 0.091 & 53 & 0.009 &1.21\\
        & \textit{black} & $>$500 & $\sim 0$ & $>$500 & $\sim 0$ &0.00\\
        \hline
    \end{tabular}
\end{table*}

\section{Conclusion}\label{sec:conclusion}
In this paper, we revisited the framework \F that was proposed to render fairer classification models when applied to tabular data.  We extended \F in several ways.
We integrated an approach to automate the choice of the list of the most important features to be taken into account during the fairness analysis.
We also evaluated different sampling strategies for selecting instances when assessing fairness.

We also explored the adaptability of \F's framework to classification models 
on textual data. For that, we 
adapted the  notion of feature dropout to bag of words, followed by the ensemble strategy. Our empirical results showed the feasibility of this idea when rendering models fairer. Furthermore, the comparison of \F's workflow with LIME and SHAP explanations indicated that, even though SHAP provides drastic improvements in some cases, \F with LIME explanations  is more stable, and consistently results in fairer models; this is not the case when using SHAP explanations. Submodular pick gives better average results than random sampling and the best gain is obtained for Logistic regression for both LIME and SHAP (and for both SP and RS).

This contribution opens several avenues for future work. For instance, in the case of textual data it remains to automate the choice of the words that \F should take into account in the dropout approach based on bags of words. Furthermore, we are currently experimenting with more complex data types, such as graphs and other structured data.
However, to guarantee competitive results, model specific explanations, namely, for deep neural approaches, need to be considered. This constitutes a topic of ongoing research.

\bibliographystyle{IEEEtran} 
\bibliography{utils/references}
%
%
%


\end{document}